\documentclass[]{ceurart}

\usepackage{listings}
\usepackage{times}
\usepackage{latexsym}
\usepackage{comment}

\usepackage[T1]{fontenc}

\usepackage[utf8]{inputenc}

\usepackage{inconsolata}

\usepackage{graphicx}

\usepackage{dsbda-style}
\usepackage{listings}
\usepackage{graphicx}
\usepackage{subcaption}
\usepackage{booktabs}
\usepackage{xspace}
\newcommand{\CRAWLDoc}{CRAWLDoc\xspace}
\newcommand{\CRAWLDocLong}{Contextual RAnking of Web-Linked Documents\xspace}

\usepackage[nolist]{acronym}

\begin{document}

\begin{acronym}
\acro{SLM}[SLM]{Small Language Model}
\acro{LLM}[LLM]{Large Language Model}
\end{acronym}

\copyrightyear{2025}
\copyrightclause{Copyright for this paper by its authors.
  Use permitted under Creative Commons License Attribution 4.0
  International (CC BY 4.0).}

\conference{SCOLIA 2025: First International Workshop on Scholarly Information Access at ECIR 2025, April 10, 2025}

\title{CRAWLDoc: A Dataset for Robust Ranking of Bibliographic Documents}

\author[1]{Fabian Karl}[%
orcid=0009-0008-0079-5604,
email=fabian.karl@uni-ulm.de,
]
\cormark[1]
\address[1]{Universität Ulm, Germany}

\author[1]{Ansgar Scherp}[%
orcid=0000-0002-2653-9245,
email=ansgar.scherp@uni-ulm.de,
]

\cortext[1]{Corresponding author.}

\begin{abstract}
Publication databases rely on accurate metadata extraction from diverse web sources, yet variations in web layouts and data formats present challenges for metadata providers. This paper introduces \CRAWLDoc, a new method for contextual ranking of linked web documents. 
Starting with a publication's URL, such as a digital object identifier, \CRAWLDoc retrieves the landing page and all linked web resources, including PDFs, ORCID profiles, and supplementary materials. It embeds these resources, along with anchor texts and the URLs, into a unified representation.
For evaluating \CRAWLDoc, we have created a new, manually labeled dataset of $600$ publications from six top publishers in computer science.
Our method \CRAWLDoc demonstrates a robust and layout-independent ranking of relevant documents across publishers and data formats.
It lays the foundation for improved metadata extraction from web documents with various layouts and formats. 

Our source code and dataset can be accessed at \url{https://github.com/FKarl/CRAWLDoc}.
\end{abstract}


\maketitle

\section{Introduction}
\label{sec:introduction}

Databases such as 
Web~of~Science\extended{\footnote{\url{https://www.webofscience.com}}}, 
Crossref\extended{\footnote{\url{https://www.crossref.org}}}, and
DBLP\extended{\footnote{\url{https://dblp.org}}} 
are crucial academic resources of bibliographic information.
Identifying high-quality metadata sources about new publications is essential for these services. 
While there are methods and tools for extracting bibliographic metadata~\cite{GROBID,CERMINE}, these are typically restricted to a single document like a PDF.
Currently, many potential web sources and content that may contain valuable metadata are underutilized.
This is due to source heterogeneity of web layouts, document types, and formats, including full texts, publication PDFs\extended{, conference websites}, publisher landing pages, ORCIDs, and other web content.

We consider the example of DBLP, the de facto main metadata provider in computer science.
The main strategy for integrating publisher-provided metadata is to implement publisher-specific wrappers, an approach that is time-consuming and requires maintenance whenever the publisher changes its website~\cite{dblp_wrapper}.
Thus, an automated service is needed to systematically search for bibliographic metadata sources across multiple web documents.
Often, bibliographic information cannot be found on a single website, e.\,g., the publication's landing page, necessitating to harvest linked documents and identifying those relevant to the publication. 
Identifying relevant linked documents is challenging because two web documents with similar layouts and text can refer to different papers with paper-specific components like titles, authors, and affiliations.
Another challenge is the heterogeneity of web data. 
Important documents can be in HTML or other formats like PDF.
\begin{figure}[thb]
    \centering
    \includegraphics[width=0.85\linewidth]{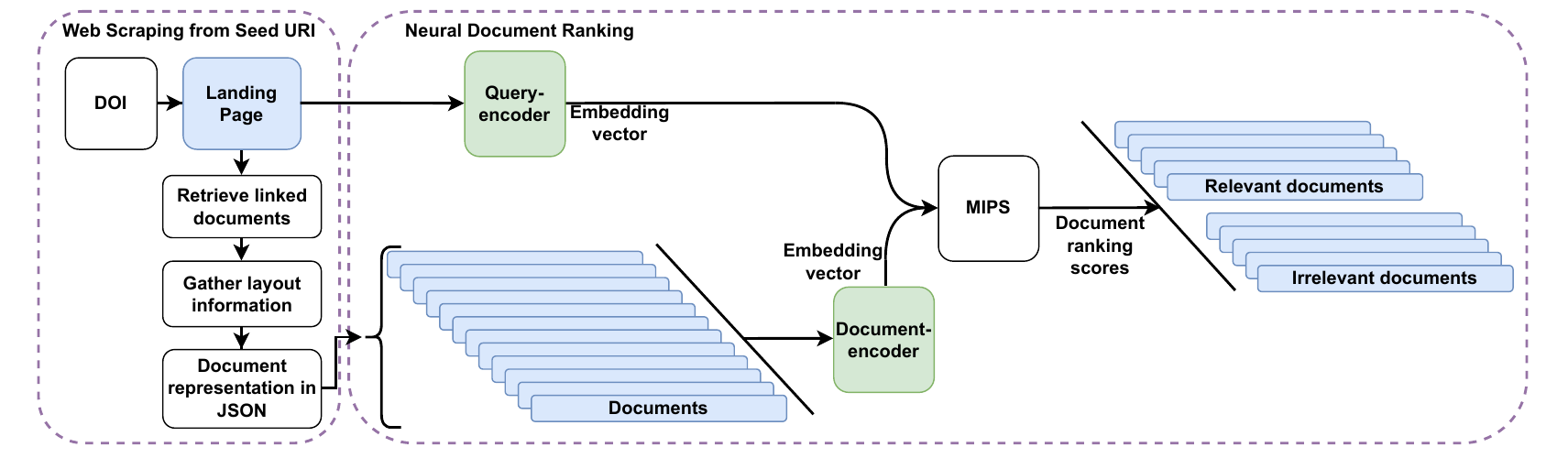}
    \caption{Illustration of document retrieval from heterogeneous web sources. 
The process starts with a DOI, which is resolved to access the landing page. 
Linked documents are ranked by treating the landing page as a query.
A maximum inner product search (MIPS) ranks the top-$k$ documents based on embeddings from a small language model.}
    \label{fig:pipeline}
\end{figure}
Another reason is that using wrappers or APIs relies on crawling publisher websites, which is expensive to maintain~\cite{dblp_wrapper}.

We propose a novel retrieval system \CRAWLDoc (\CRAWLDocLong), see Figure~\ref{fig:pipeline}, that can automatically identify relevant data sources from diverse web sources. 
Input is a Digital Object Identifier (DOI)\extended{\footnote{\url{https://www.doi.org/}}} of a publication, which is provided by publishers~\cite{DBLP_lessons_learnd}. 
The web content linked from this seed URI is harvested and analyzed.
We identify relevant linked content referring to the same paper as the DOI that may carry metadata. 
To this end, we embed the source document and linked documents along with their associated anchor texts and URLs into a shared vector space and treat the publication's landing page as the query.
A ranking is computed by the similarity between the landing page embedding and the embeddings of linked documents, effectively identifying the most relevant sources for metadata extraction.
By embedding the content, we effectively address the challenge that the web sources from which the data is extracted are highly diverse and vary in structure and format~\cite{dblp_wrapper}. 

We evaluate \CRAWLDoc on a new dataset derived from DBLP, which comprises $600$ publications from the six largest computer science publishers. 
Our dataset is unique as it provides annotated relevancy labels for all outgoing links from publication landing pages, along with bibliographic metadata, including titles, years, authors' names, and affiliations manually. 
Our experiments show that \CRAWLDoc reliably identifies relevant web documents based on a single seed document.

A leave-one-out experiment shows that our system is robust w.\,r.\,t. the retrieval from websites of various layouts from publishers that were excluded from the training dataset.
In summary, our contributions are:

\begin{itemize}
    \item %
    A document-as-query approach \CRAWLDoc to determine relevant documents 
    that encodes web content of various formats, anchor text, and URIs in a single embedding space.

    \item Evaluating 600 publications from the six largest publishers in computer science.

    \item %
    A robustness check \extended{to assess the generalizability of our system} by training on five publishers and testing on a held-out publisher.
    
    \item %
    A new dataset of bibliographic metadata with author affiliations, along with relevancy information for linked web documents.
    \extended{This dataset is the first of its kind to combine both document relevancy annotations and comprehensive bibliographic metadata. }

\end{itemize}

Below, we summarize related work.
We introduce our \CRAWLDoc metadata retrieval system in Section~\ref{sec:methods}.
The experimental apparatus is described in Section~\ref{sec:experimentalapparatus}.
The results are described in Section~\ref{sec:results} and discussed in Section~\ref{sec:discussion}, before we conclude and outline future work.

\section{Related Work}
\label{sec:relatedwork}

We discuss research in neural information retrieval and layout-aware language models.
\textbf{Neural Information Retrieval} (NIR) is a prominent research area, utilizing neural networks to improve the retrieval process.
The landscape of NIR research has been extensively surveyed~\cite{LLM_IR_survey, NIR_lit_rev, neural_ranking_survey}, highlighting the use of learned representations of queries and documents, commonly referred to as embeddings. 
These embeddings capture semantic similarities that traditional information retrieval models might overlook~\cite{nir_survey, intro_nir, nir_qa}.
\extended{
A pioneering model in this domain is the Deep Structured Semantic Model (DSSM)~\cite{dssm}. DSSM is a latent semantic model that employs a deep neural network to project queries and documents into a common low-dimensional space. In this space, the relevance of a document to a query is determined by the distance between their respective projections.}
The BERT model~\cite{bert}, although not specifically designed for information retrieval, has profoundly impacted NIR~\cite{BERT_IR_Math, BERT_IR_Survey, BERT_intra_block_ranking}. 
BERT-based models such as CEDR~\cite{cedr} have achieved impressive performance on various information retrieval benchmarks. %
The ColBERT model~\cite{colbert} introduced a late interaction paradigm, enabling efficient and effective passage retrieval. 
ColBERT's ability to balance effectiveness and efficiency has made it a popular choice for large-scale retrieval tasks~\cite{colbert_v2}.

\textbf{Layout-infused language models} consider both textual content and spatial layout. %
Lay\-out\-LMv3~\cite{LayoutLMv3} exemplifies this concept by pre-training multimodal transformers with a unified text and image masking objective\extended{, enhancing performance on both text-centric and image-centric tasks}.
Another approach is DocLLM~\cite{DocLLM}, which does not rely on expensive image encoders but relies solely on bounding box information from optical character recognition (OCR). 
\extended{This model is particularly useful for documents with irregular layouts and heterogeneous content.}
LMDX~\cite{LMDX} is a model-agnostic method to adapt arbitrary \acp{LLM} for document information extraction. 
It extracts text with OCR and enriches it with layout information\extended{ in the form of bounding boxes}. 
The model proposes an XML-style prompt for information extraction and trains a text-only \ac{LLM} with text and bounding boxes. 
\extended{The response of the \ac{LLM} is decoded as a post-processing step based on the text and bounding boxes to discard hallucinations. 
Experiments show that LMDX is effective, especially in low data regimes.}
Layout-infused \acp{LLM} can face challenges with layout distribution shifts. 
\citet{dis_shift} note that model performance can degrade by up to 20 points in macro F1 score under layout distribution shifts.

\section{\CRAWLDoc Metadata Retrieval}
\label{sec:methods}

\begin{figure}[tb]
    \centering
    \includegraphics[width=0.85\linewidth]{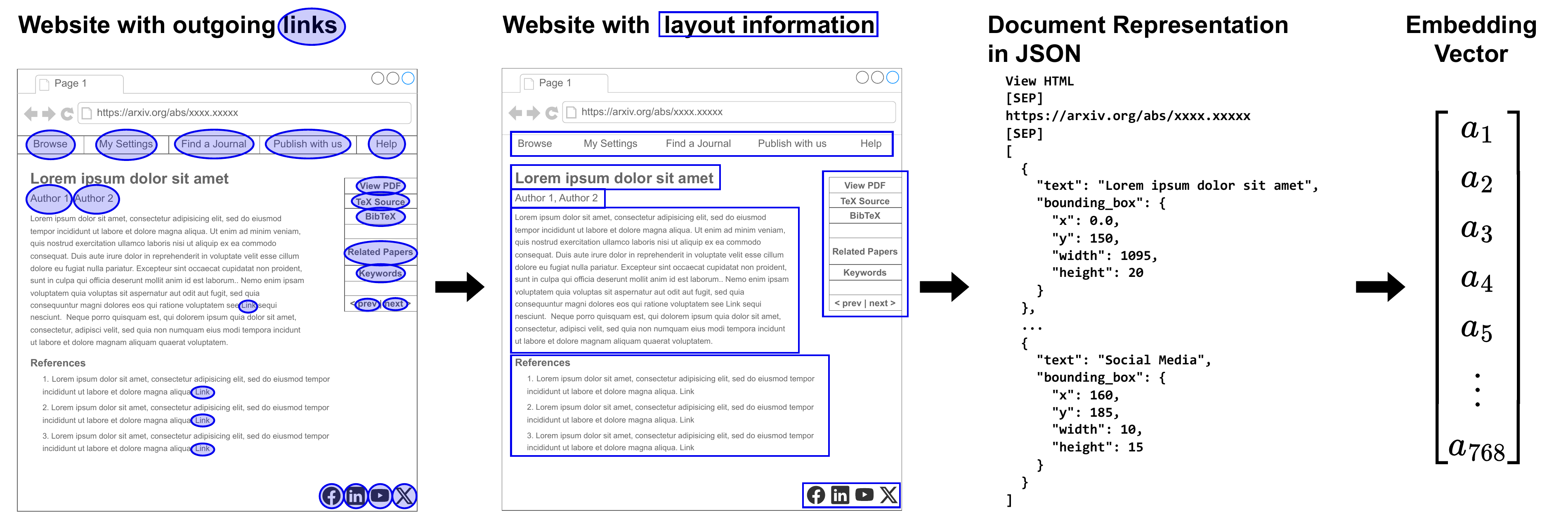}
    \caption{This figure illustrates our document representation methodology. 
    The process begins with identifying all hyperlinks on the landing page, followed by integrating layout information via bounding boxes. 
    The document is then converted into a uniform textual format and encoded into a vector representation.
    }
    \label{fig:doc_representation}
\end{figure}

We introduce \CRAWLDoc (\CRAWLDocLong), a novel system for identifying relevant bibliographic sources across web documents.
Based on a seed URI, a DOI of a publication, \CRAWLDoc scrapes linked resources.
Subsequently, the retrieved web documents in the form of HTML or PDF are ranked using a \ac{SLM}~\cite{slm_survey}.
Our primary assumption \extended{for bibliographic metadata extraction} is that all necessary information can be found within a one-hop crawl of the landing page associated with the DOI. 
This assumption is based on our observation that publishers present key bibliographic information on the landing page or pages directly linked to it \eg the PDF of the publication.

\paragraph{Web Scraping from Seed URI}
\label{sec:crawldoc:scraping}

The initial step of our system involves web scraping, starting with a DOI as the input and progressing to the scraping of the corresponding web page. 
After this starting point, all documents linked from the seed URI are retrieved, which may be formatted in HTML or PDF. 
Both PDF and HTML files undergo a series of steps to extract the relevant text and its associated bounding boxes to also capture layout information. 
For PDF documents, the text and its corresponding bounding box coordinates are directly extracted from the file using the PDFMiner Python library. 
In the case of HTML documents, the page is first rendered in a Firefox web browser (Version: 129.0.2) to accurately present the content's formatting and layout, and then the text and bounding boxes are extracted.\extended{ This ensures that the textual content and its spatial context are preserved.}
This information is then converted into a uniform textual JSON format and serves as the input for the neural document ranking.
Figure~\ref{fig:doc_representation} illustrates the different steps to create our document representation.

\paragraph{Neural Document Ranking}
\label{sec:crawldoc:ranking}

In the second step, we employ a \ac{SLM} to create \extended{unified} embeddings of the documents along with their associated anchor texts and URLs. For each document, we construct a single input representation by concatenating the anchor text, URL, and document content using a special separator token ([SEP]).
This representation is then embedded into a dense vector space. 
The document originating from the DOI is embedded utilizing a query encoder, and all documents linked from the landing page are embedded with the document encoder. A Maximum Inner Product Search~(MIPS) is performed with the embedding of the landing page and the embeddings of all scraped documents to create a Contextual RAnking of Web-Linked Documents
(CRAWLDoc) based on the landing page. 

We use the \texttt{jina-embeddings-v2~model}~\cite{jina-2} as neural retriever.
It is based on a BERT~\cite{bert} architecture and supports the symmetric bidirectional variant of ALiBi~\cite{alibi}, allowing for a sequence length of up to 81,921 tokens. 
Due to memory restrictions, we limit our experiments to the first 2,048 tokens.
The neural retriever is trained using contrastive learning with the InfoNCE loss function~\cite{info_nce}. 

\section{Experimental Apparatus}
\label{sec:experimentalapparatus}

\paragraph{Dataset}
\label{sec:datasets}

We take a subset of bibliographies from the six largest publishers in the DBLP Computer Science Bibliography  dataset~\cite{dblp_dataset}.
The publishers represent more than 80\% of all publications listed in DBLP.  %
This ensures the dataset contains a representative set of layouts encountered in bibliographic web content.
We randomly select 100 publications for each publisher and split them into training, validation, and test sets in an 80/10/10 ratio with equal per-publisher distribution.

We obtained the metadata for each publication by manually retrieving the title, publication year, and authors' names and affiliations. 
We retrieved the landing page of each publication and labeled every outgoing link on the landing page with a binary relevancy label. 
This label indicates whether the linked website or document is about the same publication \extended{and contains relevant metadata for the publication, or not}.  
By manually creating this dataset, we ensure a high quality of the metadata and can accurately assess the document retrieval process in our proposed setup. 

To prevent artificial inflation of our performance metrics, we identified and removed any instances in our test set where the landing page contained links to itself. 
The trivial nature of calculating document similarity to itself would otherwise result in an unrepresentative boost in ranking performance.

Our dataset consists of 600 publications with detailed metadata and 72,483 linked documents with binary relevancy labels. 
Per publication, we have an average of $3.63$ (SD: $2.10$) authors, with an average of
$1.14$ (SD: $0.41$) affiliations per author.
Furthermore, there is an average of
$120.81$ (SD: $76.52$) linked websites per landing page and an average of only $5.45$ (SD: $2.99$) relevant websites per publication.
To the best of our knowledge, we are the first to release a dataset that includes author affiliations as mentioned in the publications. 
Additionally, we are the first to provide relevancy labels for linked documents in the context of publication web data.
For legal purposes, we are only able to publish the labels and not the actual websites. However, we do offer the source code\extended{\footnote{\url{https://github.com/FKarl/CRAWLDoc}}} for our procedure. 

The DBLP dataset is released under CC0 1.0 Public Domain Dedication license.
Our annotations have the same license.

\paragraph{Procedure}
\label{sec:procedure}

Our experimental procedure for document ranking involves fine-tuning a neural document retriever using contrastive loss to improve document ranking. To ensure robust performance, we evaluate the ranking capabilities on both in-distribution and out-of-distribution data.
We optimize the hyperparameters of the neural document retriever resulting in a learning rate of 3e-05, 32 accumulation steps, and patience of 5, resulting in 16 epochs.

\paragraph{Metrics}
\label{sec:measures}

To evaluate the ranking of the web documents, we employ several metrics. 
The Mean Reciprocal Rank (MRR) evaluates the effectiveness of a retrieval system by considering the rank position of the first relevant result. 
\extended{It is calculated as the average of the reciprocal ranks of the first relevant result across a set of queries. 
Formally, MRR is defined as: $\mathrm{MRR} = \frac{1}{|Q|} \sum_{i=1,\ldots, |Q|} \frac{1}{rank_i}\,$.}
The MRR focuses on the first relevant document in the ranked list, \ie it favors a relevant document in the highest position.
In contrast, Mean Average Precision (MAP) evaluates the precision of a retrieval system by averaging the precision scores at all ranks where relevant documents are found and then averaging these scores over all queries. 
Normalized Discounted Cumulative Gain (nDCG)~\cite{ndcg} measures the usefulness of a document based on its position in the result list, assuming that highly relevant documents are more useful when appearing earlier.\extended{ It is computed by normalizing the Discounted Cumulative Gain (DCG) by the ideal DCG (iDCG).}
We further calculate the precision@k, recall@k, and F1@k which measure the proportion of relevant items in the top $k$ results. 

\section{Results}
\label{sec:results}

\begin{table}[tb]
\caption{Ranking performance metrics across publishers. Values are provided for each publisher and aggregated.}
\label{tab:results}
\begin{tabular}{l|cccccc|c}
\toprule
\textbf{Publisher:} & \textbf{IEEE} & \textbf{Springer} & \textbf{Elsevier} & \textbf{ACM} & \textbf{arXiv} & \textbf{MDPI} & \textbf{All} \\
\midrule
\textbf{MRR} & 1.000 & 0.800 & 1.000 & 1.000 & 1.000 & 1.000 & 0.967 \\
\textbf{MAP} & 1.000 & 0.998 & 0.970 & 0.999 & 1.000 & 0.954 & 0.987 \\
\textbf{nDCG} & 1.000 & 0.800 & 0.985 & 1.000 & 1.000 & 0.982 & 0.961 \\
\bottomrule
\end{tabular}
\end{table}

The ranking metrics for identifying relevant linked documents are shown in Table~\ref{tab:results}.
Overall, we achieve a very high average ranking performance with MRR of $0.967$, MAP of $0.987$, and nDCG of $0.961$.
The MRR, MAP, and nDCG values exhibit a consistently high level of performance for all six publishers, except for MRR and nDCG on the Springer dataset.
The MRR for IEEE, Elsevier, ACM, arXiv, and MDPI all achieve the maximum score of $1.000$, indicating that a relevant document is always placed at the top position.
To understand the impact of layout information on ranking performance, we conducted an ablation study. 
The results without layout information showed slightly lower performance with a MRR of $0.950$, MAP of $0.976$, and nDCG of $0.952$.

We have conducted a more detailed examination of the ranking performance with different cut-off values $k$ presented in Table~\ref{tab:patk}. 
The recall increases with increasing values of $k$, reaching $0.951$ at $k=10$. 
Precision declines from $0.972$ for $k=1$ to $0.416$ for $k=10$.
The F1@k score, which combines precision and recall, reaches its highest value of $0.772$ for $k=4$ and $k=5$.

\begin{table}[t]
    \caption{Ranking performance evaluation of \CRAWLDoc at different cut-off values $k$.}
    \begin{tabular}{l|rrrrrrrrrr}
        \toprule
        \textbf{k} & \textbf{1} & \textbf{2} & \textbf{3} & \textbf{4} & \textbf{5} & \textbf{6} & \textbf{7} & \textbf{8} & \textbf{9} & \textbf{10} \\
        \midrule
        \textbf{Recall@k }   & 0.344 & 0.551 & 0.692 & 0.792 & 0.870 & 0.900 & 0.932 & 0.941 & 0.943 & 0.951 \\
        \textbf{Precision@k} & 0.972 & 0.892 & 0.822 & 0.754 & 0.698 & 0.617 & 0.557 & 0.500 & 0.455 & 0.416 \\
        \textbf{F1@k}        & 0.510 & 0.678 & 0.751 & 0.772 & 0.772 & 0.732 & 0.696 & 0.652 & 0.615 & 0.579 \\
        \bottomrule
    \end{tabular}
    \label{tab:patk}
\end{table}

We have evaluated robustness using a leave-one-out strategy, training on all but one publisher and testing on the left-out publisher.
The results of the robustness analysis are shown in Table \ref{tab:robustness-check}. 
We obtain a high performance across all publishers, with an average MRR of $0.959$, MAP of $0.968$, and nDCG of $0.961$. 
This is less than one point for MRR and nDCG and less than two points for MAP compared to using the full training dataset shown in Table~\ref{tab:results}.
The results were particularly strong for IEEE and arXiv, both achieving the maximum score of $1.000$ for all three metrics. 
However, the performance was slightly lower for Springer, consistent with the result on the full training set.

\section{Discussion}
\label{sec:discussion}

\paragraph{Document Ranking}
\extended{The outcomes of our research illustrate the effectiveness of our proposed document ranking and extraction system.}
Our system shows impressive overall ranking performance, with documents ranked at the top being mostly relevant. 
This trend is seen when examining the evaluation of ranking performance at various cutoff values. 
We notice a sharp rise in recall@$k$ for the first few documents but only minor enhancements after around five documents. 
The decline in precision@$k$ as $k$ values increase is a natural result considering that a publication has on average $5.45$ relevant documents (see Section~\ref{sec:datasets}). 
This is also reflected in the F1@$k$ score, which is peaking at $k=4$ and $k=5$.
Overall, the results show that \CRAWLDoc maintains a good balance between precision and recall with a cut-off value of $k=5$.

Upon investigating cases where the model ranked irrelevant documents higher than relevant ones, we could not identify a general error pattern. 
The errors we observed were predominantly paper-specific rather than systematic.
For example, errors occurred when the model ranked links from the references section of a paper highly or when it assigned high ranks to different chapters of the same book. 
In particular, Springer publications presented more special cases than other publishers in our dataset.

\paragraph{Robustness of Document Ranking}

Our model demonstrates strong robustness across different publishers. %
While previous research, such as \citet{dis_shift}, has identified challenges for layout-infused \acp{LLM} when dealing with layout distribution shifts, our system shows consistent performance. 
This is evidenced by nearly equivalent performance between in-distribution and out-of-distribution data, suggesting effective generalization.
Academic publishers often follow similar design patterns \extended{and conventions} for their publication pages, reducing the effective layout distribution shift between sources. 
Our robustness evaluation considered six major publishers.
The conventional nature of academic publication layouts suggests our model likely generalizes to a broader range of publishers.

\paragraph{Generalization and Threat to Validity}
\label{sec:generalization}

\begin{table}[tb]
\caption{Performance of the ranking task across all publishers in a leave-one-out test. The publisher the model is ``tested on'' is not part of the training data.}
\label{tab:robustness-check}
\begin{tabular}{l|rrrrrr|r}
\toprule
\textbf{Tested on:} & \textbf{IEEE} & \textbf{Springer} & \textbf{Elsevier} & \textbf{ACM} & \textbf{arXiv} & \textbf{MDPI} & \textbf{Average} \\
\midrule
\textbf{MRR} & 1.000 & 0.757 & 1.000 & 1.000 & 1.000 & 1.000 & 0.959 \\
\textbf{MAP} & 1.000 & 0.835 & 0.996 & 0.999 & 1.000 & 0.979 & 0.968 \\
\textbf{nDCG} & 1.000 & 0.772 & 0.999 & 1.000 & 1.000 & 0.992 & 0.961 \\
\bottomrule
\end{tabular}
\end{table}

The generalizability of our work refers to different publishers based on a leave-one-out test (Table~\ref{tab:robustness-check}) in which we test the system for publishers on which it has not been trained. 
Our robustness check demonstrates that a trained model can achieve comparably good results on out-of-distribution data within our tested scope. 
This finding suggests that our model has learned generalizable features of document relevance beyond the specific layouts and publishers in our training data.
Our approach of transforming different document formats (HTML and PDF) into a uniform textual representation enhances its potential for generalization. 
This uniformity in representation suggests applicability to other web and document-related tasks.

\label{sec:threattovalidity}
While our study provides robust results, it is important to reflect on potential threats to validity. One such threat is the limited scope of our investigation, which focuses on only six publishers, primarily from the computer science field. However, the threat is reduced as these publishers represent more than $80\%$ of computer science publications and provide various formats. 
Nevertheless, we acknowledge that true generalizability to the remaining $20\%$ of publications, which may exhibit greater variability in their document layouts and metadata presentation, remains to be thoroughly tested in future work.
An additional possible risk is the presence of recency bias in our dataset, given that most publications in DBLP are from more recent years. 
Nevertheless, we have found that older publications, including papers as far back as 1967, in our test set achieve similar performance to more recent ones, which eases this concern. 
This indicates that the performance of our model is not much influenced by the publication year.

Although there is no particular reason why other embedding models could not be used, our work does not focus on finding optimal embedding models for the retrieval tasks.
We use Jina embeddings because they are widely used and have demonstrated strong results~\cite{jina-2}.

\section{Conclusion and Future Work}
\label{sec:conclusion}

Our \CRAWLDocLong (\CRAWLDoc) retrieval system effectively identifies relevant bibliographic documents across diverse web sources. The key scientific findings include robustly identifying pertinent web documents and the system's consistent performance across publishers with varying web layouts. 
The insights presented in this study can potentially advance the management and enrichment of comprehensive bibliographic databases.

\label{sec:futurework}

Although our model's performance is already very strong, rerankers~\cite{LLM_IR_survey} could improve document ranking accuracy.
Future work could also explore alternative neural retriever setups like ColBERTv2~\cite{colbert_v2} and token-level representation of documents with MaxSim~\cite{colbert} instead of cosine similarity. 
In the next steps, we plan to run different metadata extractor components and setups on the \CRAWLDoc-ranked list of web resources.
Furthermore, we plan to evaluate \CRAWLDoc in the context of the DBLP workflow.
\extended{
Developing methods to identify important sections within documents could optimize context utilization, potentially improving the efficiency and accuracy of our extraction process. 
This could be particularly beneficial when dealing with long documents or when processing time is a constraint.
Our system can also be valuable in legal and patent search~\cite{legal_search} for retrieving and extracting precise information from diverse documents.}

\begin{acknowledgments}
We thank Florian Reitz from DBLP for valuable feedback.
The authors acknowledge support by the state of Baden-Württemberg through bwHPC.
This research is co-funded by the SmartER project (No. 515537520) of the DFG, German Research Foundation. 
\end{acknowledgments}

\section*{Declaration on Generative AI}
 During the preparation of this work, the authors used Writefull and Grammarly to check spelling and grammar. They also used Writefull and Claude~3.5 Sonnet to improve the writing style. 
 After using these tools, the authors reviewed and edited the content as needed and take full responsibility for the publication’s content. 

\bibliography{sample-ceur}

\end{document}